\documentclass[11pt]{article}

\newcommand{\cmrl}{CME-GRPO}

\usepackage{acl}

\usepackage[T1]{fontenc}
\usepackage[utf8]{inputenc}
\usepackage{times}
\usepackage{latexsym}
\usepackage{inconsolata}
\usepackage{graphicx}
\usepackage{microtype}
\usepackage{hyperref}
\usepackage{url}
\usepackage{booktabs}
\usepackage{amsmath,amssymb,amsthm}
\usepackage{amsfonts}
\usepackage{nicefrac}
\usepackage{xcolor}
\usepackage[most]{tcolorbox}
\usepackage{bm}
\usepackage{enumitem}
\usepackage{multirow}
\usepackage[acronym]{glossaries}
\usepackage{etoolbox}
\usepackage{titlesec}

\usepackage{xcolor}
\definecolor{promptbg}{HTML}{F2F2F2}
\definecolor{rulecol}{HTML}{B0B0B0}
\usepackage[most]{tcolorbox}

\usepackage{CJKutf8}

\newacronym{cmp}{CMP}{Cross-Model Perplexity}
\newacronym{cme}{CME}{Cross-Model Entropy}

\newacronym{cmrl}{CME-RL}{Cross-Model Entropy Reinforcement Learning}
\newacronym{grpo}{GRPO}{Group
Relative Policy Optimization}

\makeglossaries

\theoremstyle{remark}

\setlist[itemize]{noitemsep, topsep=0pt}

\titlespacing*{\section}{0pt}{4pt plus 1pt minus 1pt}{2pt plus 1pt minus 1pt}
\titlespacing*{\subsection}{0pt}{3pt plus 1pt minus 1pt}{1pt plus 1pt minus 1pt}
\titlespacing*{\subsubsection}{0pt}{2pt plus 1pt minus 1pt}{1pt plus 1pt minus 1pt}
\titleformat{\section}{\normalfont\large\bfseries}{\thesection}{0.5em}{}
\titleformat{\subsection}{\normalfont\normalsize\bfseries}{\thesubsection}{0.5em}{}
\titleformat{\subsubsection}{\normalfont\normalsize\bfseries}{\thesubsubsection}{0.5em}{}

\AtBeginDocument{%
  \hypersetup{hidelinks}%
  \setlength{\textfloatsep}{6pt plus 1pt minus 1pt}%
  \setlength{\floatsep}{6pt plus 1pt minus 1pt}%
  \setlength{\intextsep}{6pt plus 1pt minus 1pt}%
  \setlength{\dbltextfloatsep}{6pt plus 1pt minus 1pt}%
  \setlength{\dblfloatsep}{6pt plus 1pt minus 1pt}%
  \setlength{\abovecaptionskip}{3pt}%
  \setlength{\belowcaptionskip}{0pt}%
  \setlength{\parskip}{0pt}%
}

\usepackage[normalem]{ulem}

\usepackage{xcolor}
\definecolor{deltagreen}{HTML}{1B7F3A}
\definecolor{deltared}{HTML}{B82020}
\newcommand{\dgood}[1]{{\scriptsize\textcolor{deltagreen}{$\,\Delta{}#1$}}}
\newcommand{\dbad}[1]{{\scriptsize\textcolor{deltared}{$\,\Delta{-}#1$}}}

\usepackage{times}
\usepackage{latexsym}

\usepackage[T1]{fontenc}

\usepackage[utf8]{inputenc}

\usepackage{microtype}

\usepackage{inconsolata}

\usepackage{graphicx}

\title{Label-Free Reinforcement Learning via Cross-Model Entropy}

\author{Matt Gorbett \\
  Independent Researcher \\
  \texttt{matthewgorbett@gmail.com} \\\And
  Hossein Shirazi \\
  San Diego State University \\
  \texttt{hshirazi@sdsu.edu} \\}

\PassOptionsToPackage{final}{microtype}
\begin{document}

\setlength{\textfloatsep}{6pt plus 1pt minus 1pt}
\setlength{\floatsep}{6pt plus 1pt minus 1pt}
\setlength{\intextsep}{6pt plus 1pt minus 1pt}
\setlength{\dbltextfloatsep}{6pt plus 1pt minus 1pt}
\setlength{\dblfloatsep}{6pt plus 1pt minus 1pt}
\setlength{\abovecaptionskip}{3pt}
\setlength{\belowcaptionskip}{0pt}

\maketitle

\vspace{-1em}

\begin{abstract}
Post-training large language models with reinforcement learning is bottlenecked by the reward signal. Existing approaches require either ground-truth verifiable rewards, restricting training to domains with automatic correctness checks (e.g., mathematics, code execution), or human preference labels, which are expensive to collect and prone to reward hacking. Recent label-free methods replace ground-truth verifiers with self-referential signals like majority voting or token entropy over a model's own outputs, but risk reinforcing a model's own errors. In this work we propose \gls{cme}, the mean log-likelihood of a generator's response under a \textit{separate} verifier model, as a label-free reward signal for RL post-training. \gls{cme} is continuous, training-free, and grounded in the principle that responses a verifier finds unsurprising are likely correct or high quality. Because the verifier is independent of the generator, the signal cannot be gamed through self-consistency. We integrate \gls{cme} into GRPO with no other changes to the training loop, extending label-free RL to open-ended instruction following—a regime where self-referential signals are inapplicable or poorly suited. On open-ended instruction following (UltraFeedback prompts, evaluated on AlpacaEval~2.0), \gls{cme} rewards beat the untrained base in head-to-head LLM-as-Judge comparisons across four model families (Qwen, Llama, Gemma, OLMo) and three training regimes (pretrained, SFT, and instruction-tuned), with tie-adjusted win rates ranging from $52.5\%$ to $71.4\%$. Code will be released upon publication.
\end{abstract}

\glsreset{cme}

\begin{figure}[t]
    \centering
    \includegraphics[width=\columnwidth]{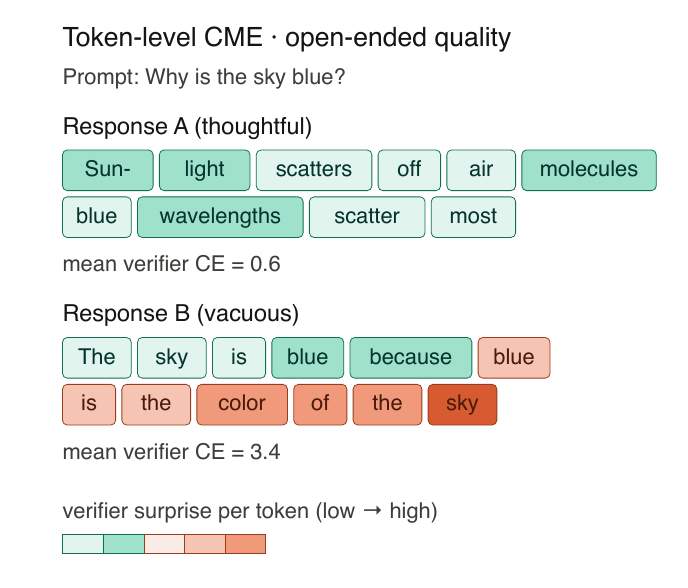}
    \caption{\textbf{Token-level CME localizes quality differences within a response.} A thoughtful answer receives uniformly low verifier surprise (mean $\text{CE}=0.6$); a vacuous one spikes as its tautology begins (mean $\text{CE}=3.4$). Per-token surprise concentrates gradient signal where responses differ in quality, without ground-truth labels.}
    \label{fig:teaser}
\end{figure}

\section{Introduction}


Reinforcement learning has emerged as a central technique for improving the reasoning and alignment of large language models. Methods like PPO~\citep{schulman2017ppo} and GRPO~\citep{shao2024deepseekmath} have driven gains on mathematical reasoning, code generation, and instruction following~\cite{guo2025deepseekr1}. These approaches rely on ground-truth verifiers, which restrict training to domains where correctness is automatically checkable. The alternative is reward models trained from human preference data~\cite{ouyang2022instructgpt}, with variants like DPO~\cite{rafailov2023dpo} eliminating the reward model and RLAIF~\cite{bai2022constitutional,lee2023rlaif} substituting an LLM judge for human annotators. However, each of these methods requires preference data, which is expensive to collect and prone to reward hacking~\cite{skalse2022reward_misspecification}.

Recent work has explored label-free alternatives. TTRL~\cite{zuo2025ttrl} and Evol-RL~\cite{zhou2025evolrl} use majority voting over sampled responses as a pseudo-label for GRPO, with Evol-RL adding a novelty term to mitigate the ``cognitive collapse'' that majority-voting alone can induce. RL-Without-GT~\cite{xin2025rl_without_gt} substitutes surface signals (e.g. format, length) for ground-truth correctness on mathematical reasoning. RENT~\cite{prabhudesai2025rent}, EM-RL~\cite{agarwal2025em}, and INTUITOR~\cite{zhao2026intuitor} take a different route, using the generator's own token entropy or self-certainty as an intrinsic reward that pushes the policy toward sharper output distributions. While each of these methods is effective in its target setting — typically math or code with verifiable structure — they share a common property: the reward derives entirely from the generator's own outputs. We refer to such rewards as \textit{self-referential}. The risk is structural: when the model is systematically wrong in a way multiple rollouts agree on, or assigns high confidence to incorrect outputs, the reward reinforces the error rather than correcting it. Majority-voting variants additionally require a canonical answer per prompt and do not extend to open-ended generation — ruling out the quality and alignment objectives that motivate modern post-training~\cite{rafailov2023dpo,bai2022constitutional}.

We propose a different approach. Instead of asking the generator to vote on its own outputs or measure its own confidence, we ask whether a separate capable model finds the generator's response surprising. \gls{cme}, the negative log-likelihood of a generator's response under a \emph{separate} verifier model, is a continuous, token-level, training-free signal that requires no ground-truth labels. Low \gls{cme} indicates the verifier finds the response unsurprising. Maximizing expected \gls{cme} minimizes the reverse KL $D_{\mathrm{KL}}(\pi_\theta \,\|\, \pi_\phi)$ up to a generator-entropy term, a useful target when the verifier is more competent in-domain than the generator. The approach is viable because open-weight instruction-tuned models like Gemma-3~\cite{gemma3n} are capable enough to serve as meaningful evaluators of response quality, even when they are not frontier models.

We integrate \gls{cme} into GRPO with no other changes to the training loop, extending label-free RL to open-ended instruction following—the regime that motivates modern post-training and where self-referential signals are inapplicable or poorly suited. We train on a subset of UltraFeedback and evaluate AlpacaEval~2.0 win rate~\citep{dubois2024alpacaeval} using LLM-as-Judge, finding that \gls{cme} rewards beat the untrained base across four model families (Qwen, Llama, Gemma, OLMo) and three training regimes (pretrained, SFT, instruction-tuned). On SFT backbones, \gls{cme}-GRPO matches DPO trained on the same base using only UltraFeedback prompts and none of its preference annotations. We additionally show that performance scales with verifier capability, and that a randomly initialized verifier control underperforms every real-weighted verifier—confirming that the signal comes from cross-model evaluation rather than generic regularization.

\section{Method}

\paragraph{Preliminaries: GRPO}
Group Relative Policy Optimization~\citep{shao2024deepseekmath} fine-tunes a policy $\pi_\theta$ without a learned critic. Given a prompt $x$, GRPO samples $G$ responses $\{y_i\}_{i=1}^{G}$ from $\pi_\theta$, scores each with a reward $r_i$, and computes group-normalized advantages $\hat{A}_i = (r_i - \mu_r)/\sigma_r$. The policy is updated via a clipped surrogate objective with a KL penalty to a frozen reference $\pi_{\text{ref}}$; the full objective is stated in Appendix~\ref{app:cme-grpo-loss}. Standard GRPO assigns a single scalar reward per response. In prior label-free work, $r_i$ is derived from majority voting~\citep{zuo2025ttrl} or format proxies~\citep{xin2025rl_without_gt}. We instead derive $r_i$ from cross-model entropy.

\paragraph{Cross-Model Entropy Reward}
We replace GRPO's reward with a label-free signal derived from a verifier model $\pi_\phi$ drawn from a different family than the generator $\pi_\theta$. Let $y_i = (y_{i,1}, \dots, y_{i,T_i})$ be a response tokenized by the generator. The per-token cross-model entropy at position $t$ is
\begin{equation}
    \mathrm{CME}_{i,t} = -\log \pi_\phi\!\left(
        y_{i,t}^{\mathrm{aligned}} \,\big|\, x,\, y_{i,<t}^{\mathrm{aligned}}
    \right),
    \label{eq:cme-token}
\end{equation}
where $y_{i,t}^{\mathrm{aligned}}$ denotes the verifier token(s) spanning the character range of the $t$-th generator token, computed by the alignment procedure of Appendix~\ref{sec:alignment-appendix}. We negate the cross-entropy so that lower verifier surprise yields higher reward: $r_{i,t} = -\mathrm{CME}_{i,t}$. Per-position advantages $\hat{A}_{i,t} = (r_{i,t} - \mu_{r,t})/\sigma_{r,t}$ are normalized across the $G$ responses at each position, and positions with no verifier counterpart are masked from the normalization and contribute no gradient. A sequence-level scalar variant of the reward, suitable when no positional credit assignment is needed, is given in Appendix~\ref{app:cme-grpo-loss}.

\newcommand{\mdl}[1]{\texttt{#1}}

\begin{figure}
\centering
\footnotesize
\begin{tcolorbox}[
  colback=gray!12, colframe=gray!75!black, boxrule=1pt,
  arc=2pt, left=4pt, right=4pt, top=3pt, bottom=3pt,
  fontupper=\footnotesize
]
\textbf{Prompt:} \emph{Did Facebook corporation change its name?}\\[2pt]
\textcolor{red!60!black}{\textbf{OLMo-DPO:}} No, Facebook did not change its name. The company's name remains Facebook.\\
\textcolor{blue!60!black}{\textbf{\cmrl{}:}} No; the platform is now known as Meta Platforms, Inc., but continues to operate under the same brand.\\[2pt]
\vspace{-1em}
\hrule 

\vspace{4pt}
\textbf{Prompt:} \emph{What is the French phrase for ``good evening''?}\\[2pt]
\textcolor{red!60!black}{\textbf{OLMo-DPO:}} ``Bonjour'' is the standard French greeting for ``good evening.''\\
\textcolor{blue!60!black}{\textbf{\cmrl{}:}} The French phrase for ``good evening'' is ``bonsoir''.
\end{tcolorbox}
\caption{Samples vs.\ \mdl{OLMo-2-0425-1B-DPO} (full responses in App.~\ref{app:samples}).}
\label{fig:samples}
\end{figure}

\begin{table*}[t]
\centering
\small
\caption{%
Pairwise tie-adjusted win rates on AlpacaEval 2.0 prompts,
where $\text{WR}_A = (\text{wins}_A + \tfrac{1}{2}\,\text{ties}) / N$.
Each win-rate cell reports the \textbf{mean across two independent LLM judges}
(\textbf{GPT-5.2} and \textbf{Claude Sonnet 4.6}).
*Reference model is \texttt{gemma-3-1b-it} for all experiments except for  \texttt{gemma-3-1b-pt} and \texttt{Qwen2.5-0.5B-It}, where we use \texttt{Llama-3.2-1B-Instruct}).
The model listed first in each comparison is the ``$A$'' winrate in ``$A$'' vs.``$B$'';
\textbf{bold} columns mark our method (\cmrl{}).
Abbreviations in-table: \texttt{Qwen2.5-0.5B-It} = \texttt{Qwen2.5-0.5B-Instruct};
\texttt{Llama-3.2-1B-It} = \texttt{Llama-3.2-1B-Instruct};
\texttt{OLMo-1B-SFT} / \texttt{-DPO} = \texttt{OLMo-2-0425-1B-SFT} / \texttt{-DPO}.
}
\label{tab:quality}
\setlength{\tabcolsep}{1pt}
\renewcommand{\arraystretch}{1.2}
\begin{tabular}{l l l c | c c | c c c}
\toprule
& & & &\multicolumn{2}{c|}{\emph{Direct head-to-head}}
  & \multicolumn{3}{c}{\emph{Anchored vs.\ Reference* }} \\
 \cmidrule(lr){5-6} \cmidrule(lr){7-9}
\shortstack[l]{\textbf{Generator}\\~}
 & \shortstack{\textbf{Pretrained}\\\textbf{baseline}}
 & \shortstack{\textbf{Instruct}\\\textbf{baseline}}
 & \shortstack{Instruct vs.\\Pretrained}
 & \shortstack{\textbf{\cmrl{}}\\\textbf{vs.\ Base}}
 & \shortstack{\textbf{\cmrl{}}\\\textbf{vs.\ Instruct}}
 & \shortstack{Base\\vs.\ Ref}
 & \shortstack{\textbf{\cmrl{}}\\\textbf{vs.\ Ref}}
 & \shortstack{Instruct\\vs.\ Ref} \\
\midrule
\multicolumn{9}{l}{\footnotesize\textit{\uline{Pretrained baselines}}} \\
\mdl{Qwen2.5-0.5B}     & \mdl{Qwen2.5-0.5B}  & \mdl{Qwen2.5-0.5B-It} & 23.0 & \textbf{71.4}\dgood{9.0} & \textbf{34.0} & 7.0  & \textbf{13.2\,\dgood{6.2}} & 22.6 \\
\mdl{Llama-3.2-1B}     & \mdl{Llama-3.2-1B}  & \mdl{Llama-3.2-1B-It} & 7.2 & \textbf{55.0}\dbad{0.7} & \textbf{6.5}  & 6.5  & \textbf{9.8\,\dgood{3.3}}  & 42.2 \\
\mdl{Gemma-3-1B-pt}    & \mdl{gemma-3-1b-pt} & \mdl{gemma-3-1b-it}   & 4.0 & \textbf{61.2}\dgood{5.2} & \textbf{9.2}  & 5.2  & \textbf{9.2\,\dgood{4.0}}  & 59.2 \\
\midrule
\multicolumn{9}{l}{\footnotesize\textit{\uline{SFT baselines}}} \\
\mdl{OLMo-1B-SFT}      & \mdl{OLMo-1B-SFT}   & \mdl{OLMo-1B-DPO}     & 40.8 & \textbf{61.4}\dgood{3.2} & \textbf{44.0} & 31.0 & \textbf{36.2\,\dgood{5.2}} & 37.6 \\
\midrule
\multicolumn{9}{l}{\footnotesize\textit{\uline{Instruction-Tuned baselines}}} \\
\mdl{Llama-3.2-1B-It}  & \mdl{Llama-3.2-1B}  & \mdl{Llama-3.2-1B-It} & 92.8 & \textbf{94.8}\dgood{2.0} & \textbf{53.3} & 7.5  & \textbf{44.2\,\dgood{2.0}} & 42.2 \\
\mdl{Qwen2.5-0.5B-It}  & \mdl{Qwen2.5-0.5B}  & \mdl{Qwen2.5-0.5B-It} & 77.0 & \textbf{83.5}\dgood{6.5} & \textbf{52.5} & 7.8  & \textbf{21.7\,\dgood{1.5}} & 20.2 \\
\bottomrule
\end{tabular}
\end{table*}

\paragraph{Character-level alignment.} When generator and verifier tokenizers differ (e.g., Qwen vs.\ Gemma), we compute $y_{i,t}^{\mathrm{aligned}}$ by summing overlapping verifier-token log-probabilities weighted by character-overlap fraction $w_{t,s} = |\mathrm{span}(v_s) \cap [a_t, b_t)| / |\mathrm{span}(v_s)|$. The weights satisfy $\sum_t w_{t,s} = 1$, preserving the total verifier log-likelihood. See Appendix~\ref{sec:alignment-appendix} for edge cases and a worked example.

\section{Experiments}
\label{experiments}

We fine-tune generators spanning four families (Qwen, Llama, Gemma, OLMo) and three training regimes (pretrained, SFT, and instruction-tuned), in each case using \texttt{gemma-3n-E4B-it}~\citep{gemma3n} as a frozen verifier. Training uses a subset of UltraFeedback~\citep{cui2024ultrafeedback} with all preference annotations and chosen/rejected completions discarded; only the prompts are retained. The reward is the token-level \gls{cme} from Eq.~(\ref{eq:cme-token}) computed over the full response, with character-level alignment to handle the cross-family tokenizer mismatch. This setup stresses the cross-family setting: the verifier shares no family with most generators, and for pretrained backbones it must provide useful gradient signal to models that have never been instruction-tuned. Full training details are in Appendix~\ref{open-ended-appendix}.


\paragraph{Evaluation.}
We evaluate on the 805-prompt AlpacaEval 2.0 set~\citep{dubois2024alpacaeval}, replacing the default GPT-4 reference with \texttt{gemma-3-1b-it} — a 1B instruction-tuned model comparable in size to the models under test. For each experiment we draw greedy responses from up to four models: the base checkpoint, our \cmrl{} fine-tune, the same-family instruct checkpoint, and the \texttt{gemma-3-1b-it} reference. We run pairwise GPT-5.2 and Claude Sonnet 4.6 LLM-as-Judge~\citep{zheng2023judging} evaluations per prompt with randomized presentation order, and report tie-adjusted win rate $(\text{wins} + \tfrac{1}{2}\text{ties})/N$ over $N=805$ prompts. Direct head-to-head comparisons (Base vs.\ \cmrl{}, \cmrl{} vs.\ Instruct) isolate \cmrl{}'s gain over same-family checkpoints; vs.-reference comparisons place all three policies on a common quality axis.

\vspace{-0.5em}

\paragraph{Results.}
Table~\ref{tab:quality} reports tie-adjusted pairwise win rates across all generators and both judges. \cmrl{} beats its untrained base in every head-to-head comparison, with margins from $52.5\%$ to $71.4\%$, and on instruction-tuned backbones it surpasses the original Instruct release on the same reference axis. Most notably, \cmrl{} matches DPO trained on the same SFT base, winning $44.0\%$ against \texttt{OLMo-2-0425-1B-DPO} using only UltraFeedback prompts and none of its preference annotations. Cross-judge agreement is high across all comparisons. Figure~\ref{fig:samples} shows representative samples, where \cmrl{} produces better prose than DPO-tuned OLMo.

\paragraph{Comparison to self-referential rewards.}
Self-referential methods like RENT \cite{prabhudesai2025rent} reward
generator confidence rather than external quality. We train RENT on
three generators with the same settings
as \cmrl{}. Table~\ref{tab:rent} shows \cmrl{} wins in every comparison,
with the largest gap on the pretrained Qwen backbone vs.\ Instruct
($34.0\%$ vs.\ $16.0\%$). As a complementary diagnostic, we use
\texttt{Qwen2.5-0.5B} as its own verifier, showing that it improves over the base
($63.0\%$ / $25.2\%$ vs.\ base/instruct) but trails cross-family \gls{cme}
with \texttt{gemma-4-E4B-it} ($70.0\%$ / $34.0\%$). A cross-family
verifier scores against an independent distribution; a same-family
verifier or self-confidence reward can only re-weight what the
generator already finds likely.

\begin{table}[t]
\centering
\small
\caption{%
\cmrl{} vs.\ RENT \cite{prabhudesai2025rent}.
Tie-adjusted win rate against each comparator (same protocol as Table~\ref{tab:quality});
\textbf{bold} marks the winner.
}
\label{tab:rent}
\setlength{\tabcolsep}{2pt}
\renewcommand{\arraystretch}{1}
\begin{tabular}{l c c | c c}
\toprule
& \multicolumn{2}{c|}{\emph{vs.\ Base}} & \multicolumn{2}{c}{\emph{vs.\ Instruct}} \\
\cmidrule(lr){2-3} \cmidrule(lr){4-5}
\footnotesize\textbf{Generator} & {\scriptsize\textbf{CME-GRPO}} &  \scriptsize\textbf{RENT} & \textbf{\scriptsize\textbf{CME-GRPO} } &  \scriptsize\textbf{RENT} \\
\midrule
{\footnotesize\mdl{Qwen2.5-0.5B}}    & \textbf{71.4} & 58.0 & \textbf{34.0} & 16.0 \\
{\footnotesize\mdl{Llama-3.2-1B-It}} & \textbf{94.8} & 67.0 & \textbf{53.3} & 50.5 \\
{\footnotesize\mdl{Qwen2.5-0.5B-It}} & \textbf{83.5} & 64.0 & \textbf{52.5} & 51.7 \\
\bottomrule
\end{tabular}
\end{table}

\section{Does Verifier Capability Matter?}
\label{sec:cme-valid}

\cmrl{} treats the verifier as a stand-in for a quality oracle, so a natural question is whether the verifier's own capability matters. We test this directly by holding the generator fixed (\texttt{Qwen2.5-0.5B}) and varying the verifier across eight conditions: a randomly initialized control, same-family verifiers (\texttt{Qwen2.5-0.5B}, \texttt{Qwen2.5-0.5B-Instruct}, \texttt{Qwen2.5-1.5B-Instruct}), and cross-family verifiers spanning 270M to 4B parameters.

\vspace{-.5em}

\paragraph{What the objective predicts.} Maximizing expected \gls{cme} minimizes the reverse KL $D_{\mathrm{KL}}(\pi_\theta \,\|\, \pi_\phi)$ up to a generator-entropy term (Appendix~\ref{app:objective}), so the natural reading is that stronger verifiers — being closer to the high-quality response distribution — should provide a more useful target. The token-level variant of Eq.~(\ref{eq:cme-token}), with character-aligned weights and group-normalized per-position advantages, is a heuristic extension that targets the same direction in distribution space without admitting the same closed form. We test this prediction empirically.

\begin{figure}[t]
\centering
\includegraphics[width=1\linewidth]{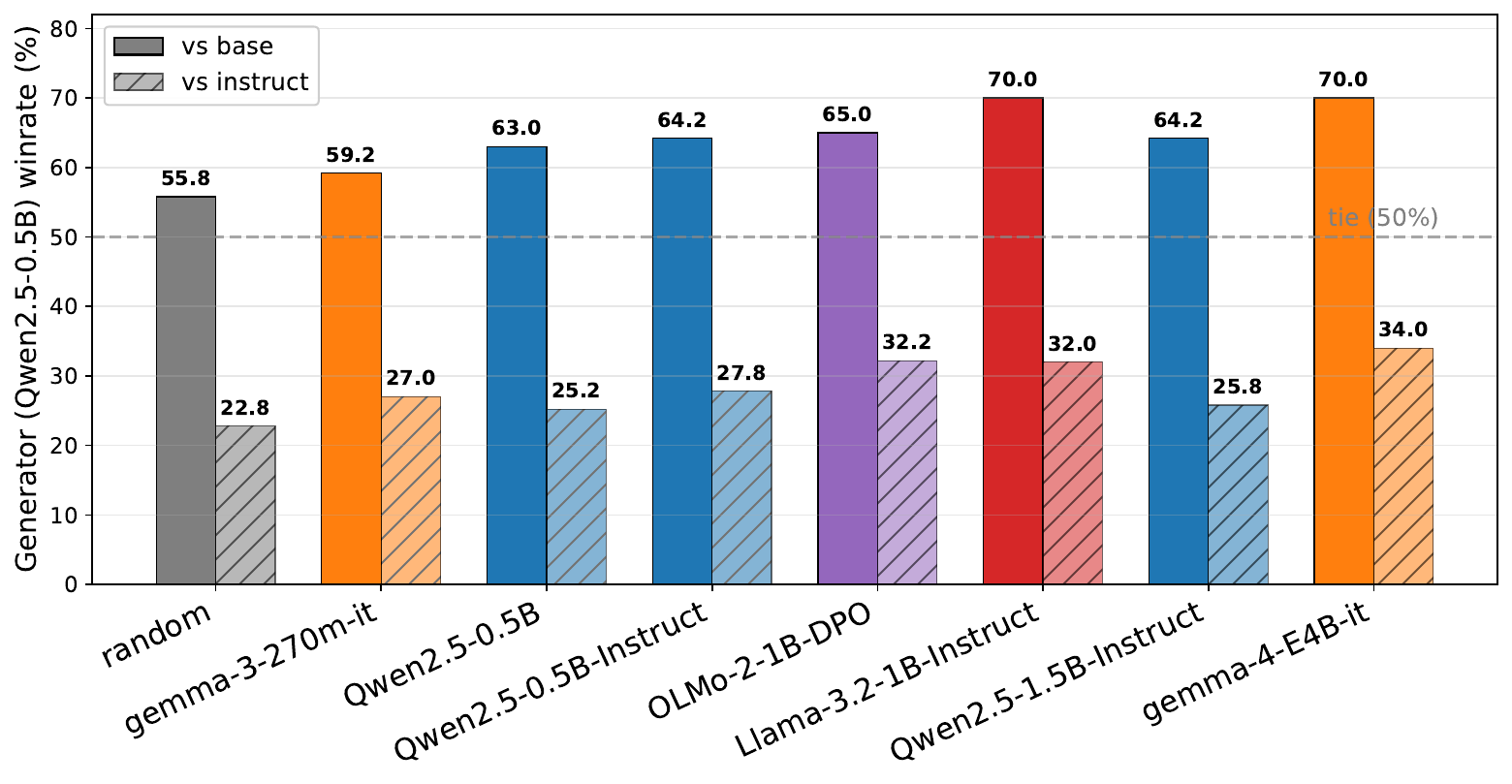}
\caption{\cmrl{} (\texttt{Qwen2.5-0.5B}) win rates on AlpacaEval 2.0 across eight verifier conditions.}
\label{fig:verifier_quality}
\end{figure}

Figure~\ref{fig:verifier_quality} reports \cmrl{} win rates against the base and instruct comparators. Three effects emerge. \textbf{(i)} The random-weighted control underperforms every real-weighted verifier ($55.8\%$ vs base, $22.8\%$ vs instruct; real-weighted verifiers exceed $59.2\%$ and $25.2\%$ respectively), confirming that CME's signal comes from cross-model evaluation rather than generic regularization. \textbf{(ii)} Among real verifiers, win rate increases with verifier capability: the largest verifier (\texttt{gemma-4-E4B-it}) ties for best vs base ($70.0\%$) and is best vs instruct ($34.0\%$), while the smallest (\texttt{gemma-3-270m-it}) sits at the bottom. The trend is not strictly monotonic in parameter count, but win rate generally increases with verifier size. \textbf{(iii)} Same-family verifiers underperform the best cross-family ones, and scaling within the Qwen family (\texttt{Qwen2.5-1.5B-Instruct}) yields no additional gain — capability matters, but family alignment does not substitute for it.

\vspace{-.5em}

\paragraph{Relation to spurious rewards.} Concurrent work~\cite{shao2025spurious} shows that RL fine-tuning of Qwen-Math models can improve from random reward signals, suggesting some label-free gains reflect surface properties rather than learned signal. Our random-verifier control quantifies this effect in the open-ended setting: a random verifier does provide some lift over the base ($55.8\%$ vs $50\%$), but every real-weighted verifier exceeds it, and gains scale with verifier capability. 

\section{Conclusion}
We introduced CME, a label-free reward for RL post-training defined as the negative log-likelihood of a generator's response under a separate verifier model. CME is a drop-in replacement for the reward in GRPO and extends label-free RL to open-ended instruction following. Across four model families and three training regimes, CME-GRPO beats the untrained base in every head-to-head comparison and matches DPO on SFT backbones using only prompts. Performance scales with verifier capability, and our random-verifier control confirms the signal comes from cross-model evaluation rather than generic regularization.

\paragraph{Limitations.} CME requires a verifier held in memory alongside the generator, adding compute and memory overhead. Because the reward measures how unsurprising a response is to the verifier rather than whether it is correct, CME is in principle vulnerable to drift toward verifier-style outputs or mode collapse toward the verifier's prior; we did not observe either failure in our experiments, but we did not test long-horizon training where these effects might emerge. Our experiments cover only small generators (0.5B–1.5B parameters) and English-language UltraFeedback prompts; behavior at larger scale or in other languages remains to be evaluated. We do not run self-entropy GRPO baselines (RENT, EM-RL) directly; our same-family verifier comparison in Section~\ref{experiments} is the closest mechanically applicable analog. Finally, we rely on LLM-as-Judge for evaluation rather than human ratings, which introduces judge-specific biases despite our use of two independent judges with high agreement.

\bibliography{custom}

\clearpage
\appendix
\onecolumn
\section*{Appendix}

\section{Full \gls{cmrl} Training Objective}
\label{app:cme-grpo-loss}

\begin{figure*}[h]
    \centering
    \includegraphics[width=\linewidth]{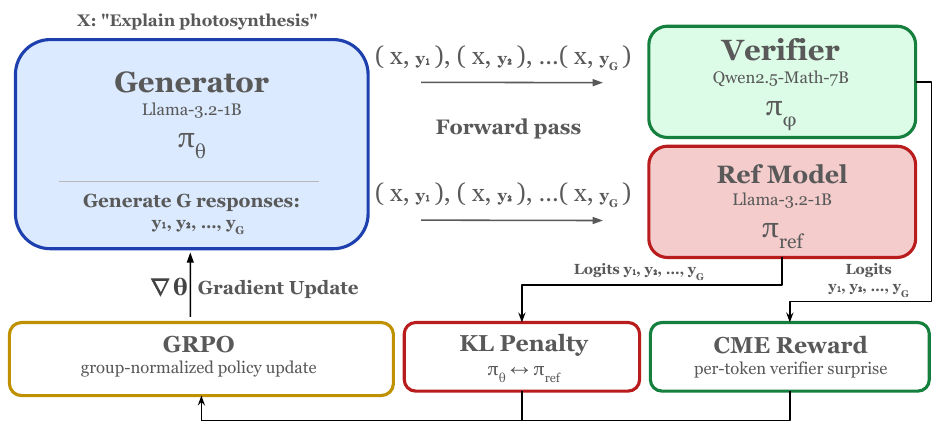}
    \caption{\textbf{Overview of CME-GRPO.} The generator $\pi_\theta$ samples $G$ responses to a prompt $x$. Each response is scored token-by-token by a verifier $\pi_\phi$, producing per-token rewards that measure verifier surprise. Group-normalized advantages are combined with a KL penalty against a reference policy $\pi_{\text{ref}}$ to form the GRPO objective.}
    \label{fig:cme-grpo}
\end{figure*}

For each prompt $x$, the generator $\pi_\theta$ samples $G$ responses $\{y_1, \ldots, y_G\}$. Each response is scored with the per-token \gls{cme} reward $r_{i,t} = -\mathrm{CME}_{i,t}$ from Eq.~(\ref{eq:cme-token}), and advantages are computed by per-position group normalization $\hat{A}_{i,t} = (r_{i,t} - \mu_{r,t})/\sigma_{r,t}$. Let
\begin{equation*}
    \rho_{i,t} = \frac{\pi_\theta(y_{i,t} \mid x, y_{i,<t})}{\pi_{\theta_{\text{old}}}(y_{i,t} \mid x, y_{i,<t})}
\end{equation*}
denote the importance-sampling ratio and $\epsilon > 0$ the PPO clipping parameter. The full \gls{cmrl} objective is
\begin{equation}
    \mathcal{L}_{\mathrm{\gls{cmrl}}} = -\frac{1}{G}\sum_{i=1}^{G} \frac{1}{|y_i|} \sum_{t=1}^{|y_i|} \min\!\Big(\rho_{i,t}\,\hat{A}_{i,t},\; \mathrm{clip}(\rho_{i,t},\,1-\epsilon,\,1+\epsilon)\,\hat{A}_{i,t}\Big) + \beta\, D_{\mathrm{KL}}\!\left(\pi_\theta \,\Big\|\, \pi_{\text{ref}}\right).
\end{equation}
This is the standard GRPO objective with reward $r_{i,t}$ replaced by the negative CME; all other terms are unchanged.

\paragraph{Sequence-level variant.} When positional credit assignment is unnecessary, we aggregate to a scalar reward $r_i = -\mathrm{CME}(y_i)$ where
\begin{equation}
    \mathrm{CME}(y_i) = \frac{1}{S_i} \sum_{s=1}^{S_i} -\log \pi_\phi(z_{i,s} \mid x, z_{i,<s})
\end{equation}
and $(z_{i,1}, \dots, z_{i,S_i})$ is $y_i$ tokenized by $\pi_\phi$. Since the sum is computed entirely within the verifier's tokenization, no cross-tokenizer alignment is required. The sequence-level reward replaces $\hat{A}_{i,t}$ with $\hat{A}_i$ in the objective above and is more efficient when quality is distributed across the full response.
\section{Character-Level Token Alignment}
\label{sec:alignment-appendix}

When the generator $\pi_\theta$ and verifier $\pi_\phi$ use different tokenizers (e.g., Qwen vs.\ Gemma), token boundaries do not align directly, so token-level CME requires mapping verifier log-probabilities onto generator positions. We decode both token sequences to a shared character string, and for each generator token $y_{i,t}$ covering character span $[a_t, b_t)$, identify the set $\mathcal{V}_t$ of verifier tokens whose spans overlap $[a_t, b_t)$. The aligned verifier log-probability at generator position $t$ is
\begin{equation*}
    \log \pi_\phi(y^{\text{aligned}}_{i,t} \mid x, y^{\text{aligned}}_{i,<t}) = \sum_{s \in \mathcal{V}_t} w_{t,s} \cdot \log \pi_\phi(v_s \mid x, v_{<s}),
\end{equation*}
where $w_{t,s} = |\mathrm{span}(v_s) \cap [a_t, b_t)| / |\mathrm{span}(v_s)|$ distributes a verifier token's log-probability across generator positions in proportion to character overlap. The weights satisfy $\sum_t w_{t,s} = 1$ for every verifier token, so per-position rewards aggregate to the correct total verifier log-likelihood of the response regardless of tokenizer mismatch.

\paragraph{Edge cases.} When generator and verifier tokens cover identical spans, $\mathcal{V}_t$ contains a single verifier token with weight one. When the generator tokenizer is coarser, verifier tokens fall fully inside generator tokens and contribute their full log-probability (the unweighted sum equals the joint character-span log-probability by the chain rule). When the verifier tokenizer is coarser, fractional weighting prevents the verifier token's log-probability from being counted multiple times. Generator tokens that decode to whitespace or have no overlapping verifier token are assigned zero aligned log-probability and masked from advantage normalization.

\paragraph{Worked example.} For the response \texttt{"unhappiness"} with generator tokenization \texttt{["un", "happiness"]} (spans $[0,2)$, $[2,11)$) and verifier tokenization \texttt{["unhappy", "ness"]} (spans $[0,7)$, $[7,11)$), the verifier token \texttt{"unhappy"} straddles both generator tokens: $2/7$ of its log-probability falls in \texttt{"un"} and $5/7$ in \texttt{"happiness"}. The verifier token \texttt{"ness"} sits entirely within \texttt{"happiness"} and contributes its full log-probability. The two aligned per-position log-probabilities sum to the total verifier log-likelihood of \texttt{"unhappiness"}.

\paragraph{Why not average?} Averaging verifier log-probabilities over $\mathcal{V}_t$ would make the aggregate reward depend on tokenization granularity rather than response content: a generator token covering $k$ verifier tokens would receive $1/k$ of their summed log-probability rather than the full sum. This would also underweight spans the verifier tokenizes finely — exactly the spans where verifier surprise is most informative.

\section{Reverse-KL Connection}
\label{app:objective}

The \gls{cme} reward connects to reverse-KL distillation through the cross-entropy decomposition. For the sequence-level reward $r(y) = \log \pi_\phi(y \mid x)$,
\begin{equation*}
\begin{split}
    \mathbb{E}_{y \sim \pi_\theta}[r(y)]
    &= -H(\pi_\theta, \pi_\phi) \\
    &= -H(\pi_\theta) - D_{\mathrm{KL}}(\pi_\theta \,\|\, \pi_\phi),
\end{split}
\end{equation*}
so maximizing expected \gls{cme} pulls $\pi_\theta$ toward $\pi_\phi$. This identity is exact for the sequence-level reward under the verifier's tokenization; the token-level variant of Eq.~\ref{eq:cme-token}, with character-aligned weights and group-normalized per-position advantages, is a heuristic extension that targets the same direction in distribution space without admitting the same closed form.

\gls{cme}-GRPO is thus closely related to MiniLLM~\citep{gu2024minillm}, which optimizes reverse KL between student and teacher via policy gradient on student samples scored by the teacher. The differences are operational: GRPO's group-normalized advantages depend only on the relative ranking of \gls{cme} values within a sampled group, and a KL anchor to the reference policy bounds drift toward the verifier. We therefore view \gls{cme}-GRPO as an RL formulation of on-policy reverse-KL distillation, with the verifier acting as a critic rather than a target distribution.

\section{Related Work}
\label{app:extended_related_work}

\subsection{RLHF and Preference Optimization}

The modern RLHF pipeline was formalized by \citet{ouyang2022instructgpt} in InstructGPT, which trained GPT-3 to follow instructions using human preference labels and PPO~\cite{schulman2017ppo}. This established the three-stage recipe---SFT, reward modeling, and RL---that underpinned the first generation of aligned language models.

A key limitation of RLHF is the cost of human preference data. \citet{bai2022constitutional} introduced Constitutional AI and RLAIF, using an LLM to generate preference labels from written principles. \citet{lee2023rlaif} compared RLAIF to RLHF and found comparable alignment quality, validating AI feedback as a substitute. CME goes further: rather than prompting a model to judge quality, we derive a reward directly from token-level probabilities, requiring no prompting or structured judgment.

DPO~\cite{rafailov2023dpo} eliminated the reward model by re-parameterizing the RLHF objective as a classification loss on preference pairs. DPO is simpler than PPO-based RLHF but requires offline preference pairs. Online variants such as Online DPO and IPO~\cite{azar2024ipo} address distribution shift by generating fresh completions during training. CME-GRPO is fully online and label-free.

\subsection{Reinforcement Learning with Verifiable Rewards}

The insight enabling GRPO~\cite{shao2024deepseekmath} was that mathematical correctness provides a free, scalable reward: reward 1 if the final answer is correct, else 0. DeepSeek-R1~\cite{guo2025deepseekr1} scaled this approach and showed emergent reasoning behaviors including self-reflection without supervised reasoning traces. Subsequent improvements include DAPO~\cite{yu2025dapo}, which introduced decoupled clipping and dynamic sampling for large-scale GRPO stability, and PRIME~\cite{cui2025prime}, which augments GRPO with process-level reward shaping. All operate in verifiable domains; we extend GRPO to non-verifiable domains by substituting cross-model entropy for the ground-truth verifier.

\subsection{Self-Improvement Without Labels}

STaR~\cite{zelikman2022star} bootstrapped reasoning by iteratively fine-tuning on correct self-generated rationales, requiring only binary answer verification. Self-consistency~\cite{wang2022self_consistency} showed that majority voting over reasoning paths improves accuracy at inference time. TTRL~\cite{zuo2025ttrl} converted self-consistency into a training signal, using majority vote as a pseudo-label for GRPO. RL-Without-GT~\cite{xin2025rl_without_gt} showed that format and response length serve as effective surrogate signals for math training. Evol-RL~\cite{zhou2025evolrl} identified ``cognitive collapse'' in majority-voting methods---models losing solution diversity---and added a semantic novelty reward. MM-UPT~\cite{wei2025mmupt} extended unsupervised GRPO to multimodal LLMs.

All of these methods are fundamentally self-referential: the reward derives entirely from the model's own outputs. If the model is systematically wrong in a way multiple rollouts agree on, the reward reinforces the error. CME-GRPO addresses this by drawing the reward from an external model in a different family.

\subsection{Knowledge Distillation.} Maximizing expected \gls{cme} minimizes $D_{\mathrm{KL}}(\pi_\theta \,\|\, \pi_\phi)$ on-policy, the reverse-KL objective optimized by MiniLLM~\citep{gu2024minillm,hinton2015distilling}. \gls{cme}-GRPO differs operationally: group-normalized advantages depend only on the relative ranking of \gls{cme} values within a sampled group, and a KL anchor to the reference policy bounds drift toward the verifier — yielding an RL formulation in which the verifier acts as a critic rather than a target distribution.

\subsection{Cross-Model Likelihood as a Correctness Signal}
\label{app:cross_model_signals}

Prior work~\citep{gorbett2026cmp} establishes that verifier likelihood ranks correct above incorrect responses across diverse tasks and model families, framed as a passive inference-time correctness signal for evaluation and selection. Our work translates this passive signal into an active training objective: rather than scoring fixed responses, we use verifier likelihood as the reward in policy optimization, and introduce a token-level variant with dense per-position advantages. 

\subsection{Process Reward Models}

Outcome reward models (ORMs) assign a single scalar to a complete response, creating sparse credit assignment. \citet{lightman2023lets} addressed this with process reward models (PRMs) trained on human-labeled step-level correctness (PRM800K), which consistently outperform ORMs on reasoning tasks. The bottleneck is annotation cost. Math-Shepherd~\cite{wang2024mathshepherd} automated PRM supervision via Monte Carlo estimation. \citet{luo2024automated_process} scaled this further, and \citet{zheng2025prm_lessons} showed LLM-as-a-judge outperforms MC estimation for PRM data quality. \citet{prm_survey2025} surveys PRMs across math, code, multimodal, and agentic domains. The token-level CME variant provides per-token process rewards that are training-free, requiring only verifier forward passes.

\subsection{Reward Signal Reliability and Goodhart's Law}

\citet{shao2025spurious} showed that Qwen2.5-Math models improve substantially even with random, format-only, or incorrect reward signals, suggesting RLVR on these models may surface latent pretraining knowledge rather than new reasoning. The effect is model-family-specific: Llama and OLMo models do not improve with spurious rewards. This motivates evaluating CME on both Qwen and non-Qwen generators to ensure results reflect genuine signal quality. Reward hacking more broadly~\cite{skalse2022reward_misspecification,pan2022reward_effects} is mitigated in CME by the cross-family verifier design: the generator cannot exploit biases of a model from a different training lineage as easily as it can exploit its own self-consistency.

\subsection{LLM Evaluation Benchmarks}

Mathematical reasoning benchmarks used in our evaluation include the MATH dataset~\cite{hendrycks2021math}, MATH-500~\cite{lightman2023lets}, AIME 2024~\cite{aime2024}, and AMC 2023. Instruction-following quality is measured via AlpacaEval 2.0~\cite{dubois2024alpacaeval}, which reports win rate against GPT-4-turbo using an LLM judge.

\section{Experiment Details}
\label{open-ended-appendix}

\paragraph{Generators and verifier.}
We fine-tune two base generator models of comparable scale but
different families: \texttt{Qwen2.5-0.5B}~\cite{yang2024qwen25} and
\texttt{Llama-3.2-1B}~\cite{llama32}. The verifier is
\texttt{gemma-3-1b-it}~\cite{gemma3n}, drawn from a third model
family to ensure cross-family \gls{cme} signal. The verifier weights are kept
in bf16 and never updated; they incur only a single forward pass per
generated response.

\paragraph{Training data.}
We sample a 5{,}000-instruction subset of UltraFeedback~\cite{cui2024ultrafeedback},
retaining only the \texttt{instruction} field and discarding all
human preference annotations and chosen/rejected completions. The
subset is held fixed across both generator runs.

\paragraph{Reward and objective.}
We use the token-level \gls{cme} from Eq.~(\ref{eq:cme-token}) computed
over the entire generator response, with character-level alignment
between the generator's and verifier's tokenizers
(Section~\ref{sec:alignment}). Per-position advantages are normalized
across the group of $G=8$ rollouts at each token position, and the
\cmrl{} objective is applied with
KL anchor coefficient $\beta = 0$ and PPO clipping $\epsilon = 0.2$.

\paragraph{Training hyperparameters.}
Both generators are trained for one epoch on the 5K subset.
Sampling uses temperature $T=1.0$, top-$p=1$, and max response length
$2048$ tokens. We use AdamW with peak learning rate $3{\times}10^{-6}$
(Qwen) and $5{\times}10^{-6}$ (Llama), linear warmup over $20$ steps
followed by cosine decay, gradient norm clipped at $1.0$, and bf16
mixed precision. The effective batch is $128$ prompts per optimizer
update (per-device batch $1$, $G=8$ rollouts, $128/G = 16$ unique
prompts $\times$ $G$ accumulated micro-steps$\dots$ effective $128$
completions). All runs use a single NVIDIA A6000 (48GB).

\paragraph{Evaluation.}
\textit{Pairwise judge on AlpacaEval 2.0 prompts.}
We use the $805$-prompt AlpacaEval 2.0 evaluation
set~\cite{dubois2024alpacaeval} as a fixed, diverse held-out instruction
distribution. We anchor against
\texttt{gemma-3-1b-it}, a small but capable instruction-tuned
checkpoint comparable in scale to the systems under test. For each
generator backbone we generate greedy responses on the full $805$
prompts from \emph{four} models---the untrained base policy, our
\cmrl{} fine-tune, the same-family instruct checkpoint, and the
\texttt{gemma-3-1b-it} reference---and run \emph{five} pairwise
GPT-5.2 judgments per prompt with randomized $A$/$B$ presentation
order to mitigate position bias: $\textsc{Base}$ vs.\ $\textsc{Cme}$,
$\textsc{Cme}$ vs.\ $\textsc{Instruct}$, and each of the three policies
against the \texttt{gemma-3-1b-it} reference. We report tie-adjusted
win rate, $(\text{wins} + \tfrac{1}{2}\,\text{ties}) / N$, following
AlpacaEval convention. The first two comparisons isolate the gain of
\cmrl{} relative to the same-family base and instruct checkpoints
(controlled by model architecture and pretraining); the three
vs.-reference comparisons place all three policies on a common
quality axis defined by an external small instruction-tuned model.
Per-sample judgments---verdict and a one-sentence justification---are
persisted to disk for inspection.

\section{Open-Ended Instruction Samples}
\label{app:samples}

\subsection{Llama-3.2-1B-Instruct generator}
 
\begin{tcolorbox}[title=\textbf{Win}: more useful clarifying questions]
\textbf{Prompt:} Hi, I'm trying to solve a crossword puzzle, but I've never done one of these before. Can you help me out?

\itshape 
\medskip\hrule\medskip
\textbf{\textcolor{green!50!black}{\texttt{Llama-3.2-1B-Instruct} (instruct):}}\\
I'd be happy to help you solve your crossword puzzle. What clues are you having trouble with? Do you have any letters already filled in, or would you like me to generate a random clue for you to start with?
\medskip\hrule\medskip
\textbf{\textcolor{blue!70!black}{\cmrl{} (ours):}}\\
I'd be happy to help you solve your crossword puzzle. What clues are you having trouble with? Do you have any letters already filled in, or are you starting from scratch? What's the size of the puzzle, and what's the theme (if any)?
\end{tcolorbox}
 
\begin{tcolorbox}[title=\textbf{Loss}: fabricated band history]
\textbf{Prompt:} Who was the first lead vocal for the rock band Starship?
\itshape 
\medskip\hrule\medskip
\textbf{\textcolor{green!50!black}{\texttt{Llama-3.2-1B-Instruct} (instruct):}}\\
I do not have information on who the first lead vocalist for the rock band Starship was.
\medskip\hrule\medskip
\textbf{\textcolor{blue!70!black}{\cmrl{} (ours):}}\\
I couldn't find any information on a rock band called ``Starship.'' However, I did find that the band ``Starship'' was formed in 1974 and was originally called ``Starship II.'' The lead vocalist for the band was Mickey Thomas.
\end{tcolorbox}

\subsection{Qwen2.5-0.5B-Instruct generator}
 
\begin{tcolorbox}[title=\textbf{Win}: factual accuracy on medical content]
\textbf{Prompt:} How do vaccinations work to protect individuals and communities from infectious diseases, and what is herd immunity?
\itshape 
\medskip\hrule\medskip
\textbf{\textcolor{green!50!black}{\texttt{Qwen2.5-0.5B-Instruct} (instruct):}}\\
Vaccinations work in several key ways to protect individuals and communities from infectious diseases:
\begin{enumerate}\itemsep0pt
\item Immunization: Vaccines stimulate the body's immune system to produce antibodies against specific pathogens.
\item Immune memory: After receiving a vaccine, the immune system produces memory cells that can quickly recognize and respond to future infections.
\item Protection: The vaccinated individual will be protected if they contract the disease after being exposed to it through exposure to the virus or bacteria contained within the vaccine.
\item Community protection: By protecting vulnerable populations such as infants, elderly people, and those with weakened immune systems, vaccination programs help reduce the overall burden of infectious diseases on community health.
\end{enumerate}
Herd immunity refers to the phenomenon where a sufficient number of individuals in a population become immune to a particular disease through natural infection or vaccination. \textbf{For example, measles has been eradicated worldwide due to high levels of herd immunity among children who have been vaccinated against the disease. In contrast, smallpox was eradicated in 1980 because most people had died from the disease before it could spread widely enough for herd immunity to occur.}
\medskip\hrule\medskip
\textbf{\textcolor{blue!70!black}{\cmrl{} (ours):}}\\
Vaccinations work by introducing small amounts of an infectious disease-causing agent into the body through injections or other means. When the immune system recognizes the agent as foreign and mounts an immune response, it can produce antibodies that protect against future exposure to the same agent.
 
Herd immunity occurs when a large number of people in a community have been vaccinated against a particular infectious disease. When this happens, even if one person gets infected with the disease, the remaining members of the community will be protected because they have already received the vaccine and their immune system has already developed antibodies against the pathogen.
 
Herd immunity is important because it helps to reduce the spread of infectious diseases within a community. By protecting those who are not vaccinated, herd immunity can help to slow down the spread of the disease and prevent its devastating effects on human health.
\end{tcolorbox}
 
\begin{tcolorbox}[title=\textbf{Loss}: unnecessary elaboration introduces factual errors]
\textbf{Prompt:} What is the capital of Australia?

\itshape 
\medskip\hrule\medskip
\textbf{\textcolor{green!50!black}{\texttt{Qwen2.5-0.5B-Instruct} (instruct):}}\\
The capital of Australia is Canberra.
\medskip\hrule\medskip
\textbf{\textcolor{blue!70!black}{\cmrl{} (ours):}}\\
The capital of Australia is Canberra. Canberra is the largest city in Australia and serves as the capital of the country. It is located on the northern coast of the continent and is known for its beautiful scenery, rich culture, and advanced technology. The city has a population of over 3 million people and is home to many important institutions, including the Australian Parliament Building, the National Museum of Australia, and the Commonwealth Scientific and Industrial Research Organization (CSIRO). Canberra is also known for its unique architecture, such as the Parliament House and the Royal Botanic Garden.
\end{tcolorbox}

\subsection{SmolLM2-1.7B generator}
 
\begin{tcolorbox}[title=\textbf{Win}: avoids fabricating a fictional ability]
\textbf{Prompt:} Give a brief description of the book that you are given. \emph{Matilda}

\itshape 
\medskip\hrule\medskip
\textbf{\textcolor{green!50!black}{\texttt{SmolLM2-1.7B-Instruct} (instruct):}}\\
Matilda is a children's novel written by Roald Dahl, first published in 1988. The story revolves around a young, intelligent, and spirited girl named Matilda, who possesses a unique set of abilities. Matilda's life is marked by her extraordinary intelligence, her love for reading, and \textbf{her ability to communicate with animals}. She lives with her parents, the Wormwood family, who are neglectful and abusive\dots
\medskip\hrule\medskip
\textbf{\textcolor{blue!70!black}{\cmrl{} (ours):}}\\
Matilda is a children's novel written by Roald Dahl, first published in 1988. The story revolves around a young, intelligent, and spirited girl named Matilda, who possesses extraordinary abilities and a unique perspective on the world. The novel is filled with humor, adventure, and a touch of magic, as Matilda navigates her life with her supportive family and her loving aunt, Miss Honey\dots
\end{tcolorbox}
 
\begin{tcolorbox}[title=\textbf{Loss}: factual error in character roleplay]
\textbf{Prompt:} Hello there Obi One Kenobi
\medskip\hrule\medskip

\textbf{\textcolor{green!50!black}{\texttt{SmolLM2-1.7B-Instruct} (instruct):}}\\
Ah, greetings, young one. I am Obi-Wan Kenobi, a Jedi Knight. I sense that you are curious about the ways of the Force. May I ask, what brings you to seek out a Jedi such as myself?
\medskip\hrule\medskip
\textbf{\textcolor{blue!70!black}{\cmrl{} (ours):}}\\
Ah, greetings, young one. I am Obi-Wan Kenobi, \textbf{a Jedi Master from the planet of Naboo}. I sense that you are curious about the ways of the Force, and perhaps seeking guidance. What is it that you wish to know, or what is it that you seek?
\end{tcolorbox}

\subsection{OLMo-2-0425-1B-SFT generator}
 
\begin{tcolorbox}[title=\textbf{Win}: calibration on an underspecified request]
\textbf{Prompt:} can you write code?
\itshape 
\medskip\hrule\medskip
\textbf{\textcolor{green!50!black}{\texttt{OLMo-2-0425-1B-DPO} (instruct):}}\\
Certainly! Below is a simple Python code snippet that calculates the factorial of a number using recursion.
\medskip\hrule\medskip
\textbf{\textcolor{blue!70!black}{\cmrl{} (ours):}}\\
Of course! I can help you write code for various tasks. Please provide me with the specific code you need, and I'll do my best to assist you.
\end{tcolorbox}
 
\begin{tcolorbox}[title=\textbf{Loss}: arithmetic error yields impossible result]
\textbf{Prompt:} Solve for x in the equation $3x + 10 = 5(x - 2)$.
\medskip\hrule\medskip
\textbf{\textcolor{red!70!black}{\texttt{OLMo-2-0425-1B-SFT} (base):}}\\
\itshape 
\medskip\hrule\medskip
\textbf{\textcolor{green!50!black}{\texttt{OLMo-2-0425-1B-DPO} (instruct):}}\\
First, distribute the 5 on the right side of the equation: $5(x - 2) = 5x - 10$. Now, combine like terms on the left side: $3x + 10 = 5x - 10$. Next, subtract $3x$ from both sides to isolate the $x$ terms on one side: $10 = 2x - 10$. Finally, add 10 to both sides to solve for $x$: $20 = 2x$. Divide both sides by 2: $x = 10$.
\medskip\hrule\medskip
\textbf{\textcolor{blue!70!black}{\cmrl{} (ours):}}\\
To solve for $x$ in the equation $3x + 10 = 5(x - 2)$, follow these steps:
\begin{enumerate}\itemsep0pt
\item Distribute the 5 on the right side of the equation: $5(x - 2) = 5x - 10$.
\item Combine like terms on the right side: $5x - 10 = 5x - 10$.
\item Subtract $5x$ from both sides to isolate the $x$ terms: $0 = -10$.
\item \textbf{Divide both sides by $-10$ to solve for $x$: $x = 0$.}
\end{enumerate}
\end{tcolorbox}

\end{document}